# Roadside LiDAR Vehicle Detection and Tracking Using Range and Intensity Background Subtraction

**Tianya Zhang** Ph.D. [a*], **Peter J. Jin** Ph.D. [a]

[a] Department of Civil and Environmental Engineering, Rutgers, The State University of New Jersey, 500 Bartholomew Rd, Piscataway, NJ 08854-8018, United States

**ABSTRACT**

In this paper, we developed the solution of roadside LiDAR object detection using a combination of two unsupervised learning algorithms. The 3D point clouds are firstly converted into spherical coordinates and filled into the elevation-azimuth matrix using a hash function. After that, the raw LiDAR data were rearranged into new data structures to store the information of range, azimuth, and intensity. Then, the Dynamic Mode Decomposition method is applied to decompose the LiDAR data into low-rank backgrounds and sparse foregrounds based on intensity channel pattern recognition. The Coarse Fine Triangle Algorithm (CFTA) automatically finds the dividing value to separate the moving targets from static background according to range information. After intensity and range background subtraction, the foreground moving objects will be detected using a density-based detector and encoded into the state-space model for tracking. The output of the proposed solution includes vehicle trajectories that can enable many mobility and safety applications. The method was validated at both path and point levels and outperformed the state-of-the-art. In contrast to the previous methods that process directly on the scattered and discrete point clouds, the dynamic classification method can establish the less sophisticated linear relationship of the 3D measurement data, which captures the spatial-temporal structure that we often desire.

Note to Practitioners—This article addresses a critical roadblock: LiDAR data often consume too much computation and processing resources, which could be eliminated by separating LiDAR data into static backgrounds and moving foregrounds. One of the main findings from the paper is that the unstructured point clouds can be transformed into a structured format and efficiently reduce the complexity of LiDAR data processing. The outcome of this research could significantly improve LiDAR data acquisition, modeling, and storage.

∗Corresponding author
✉ tz140@scarletmail.rutgers.edu (T. Zhang); peter.j.jin@rutgers.edu (Peter J. Jin)
ORCID(s): 0000-0002-7606-9886 (T. Zhang); 0000-0002-7688-3730 (Peter J. Jin)

# INTRODUCTION

Light Detection and Ranging (LiDAR) is a high-precision sensor that uses a laser transmitter and receiver to detect the distance of the surrounding object and provides 3D information of the environment. The LiDAR sensor can meet the requirements for most scenarios, particularly suitable for moving object detection and localization. The LiDAR sensor has recently gained escalating traction for smart city and connected infrastructure applications such as intelligent intersections to ensure pedestrian and bicycle safety, parking and construction management, drone-based traffic monitoring, etc. LiDAR technology is beneficial for object motion detection, especially under some low light conditions, as it can see the surrounding environment both day and night. LiDAR sensors can provide high-resolution results while radar does not have enough resolution. Although the accuracy of the LiDAR sensor could be impacted by specific scenarios due to phantom reflections, for instance, the fog weather, the LiDAR sensor is very reliable for data collection in every lighting for the multimodal traffic monitoring system. LiDAR sensors generate data for scene depth understanding, whereas the camera-based system cannot generate precise depth estimation directly. Another advantage of the LiDAR sensor is that the 3D point cloud data do not introduce any privacy concerns, which is critical for security purposes. The detection results could be used for real-time traffic signal optimization to reduce pedestrian/cyclists' waiting time and protect vulnerable road users at the signalized intersection. With the LiDAR data collection tool, the traffic manager can learn the mobility patterns to understand the causes for non-recurrent or recurrent congestion. Connected Vehicle applications also rely on real-time data acquisition capability to enable Connected Vehicle (CV) applications through V2X (Vehicle-to-Everything) communications to address safety and mobility challenges.

The usage of LiDAR technologies has also been questioned whether the LiDAR sensor has enough value to be a good investment. LiDAR is viewed as a complementary sensor to the camera or radars for the connected infrastructure solution. Critics have argued that the LiDAR sensor is a crutch to the Vision Zero of future traffic. In general, combing different sensors will increase the reliability and benefit the analyzing layer. It also serves as a non-intrusive approach to provide sufficient coverage. If powered with the next-generation communication network, the 3D data can be accessed in real-time to enable safety-critical applications and cyber-physical modeling for computational decision-making.

The majority of the LiDAR-based object detection models were developed for self-driving applications. Current deep learning approaches for autonomous driving LiDAR application are often inadequate for roadside LiDAR because the models often make ineffective predictions on new scenarios with no training data. Infrastructure-based LiDAR has its unique characteristics, compared to mobile LiDAR or airborne LiDAR, leading to different object detection approaches. As the roadside LiDAR contains mostly static backgrounds, an efficient and robust background modeling approach is proposed in this paper. More specifically, our proposed method can automatically extract background features through intensity and range information. As an unsupervised learning method for roadside LiDAR application, this method has better explanatory capability and does not need any labeled data. The data-driven algorithm was built on a solid theoretical foundation than earlier roadside LiDAR methods that rely on observational indicators. This method required few parameters, making it suitable to be used as the benchmark for future works. After a thorough evaluation, this method has achieved the new state-of-the-art by relating the rich body of background modeling techniques to the LiDAR point clouds.

**RELATED WORK**

**Roadside LiDAR Detection**

LiDAR-based object detection algorithms are categorized as mobile LiDAR object detection methods and static LiDAR object detection methods. Most LiDAR data processing algorithms were developed to use precise 3D geometric information for autonomous driving vehicles. Commonly used data structures include point-based [1] , voxel-based [2] , pillar based [3] , project-based [4, 5, 6, 7] and graph-based [8] methods. The mobile LiDAR sensors are carried through complex environments with all captured data for scene understanding. Given the many differences between LiDAR autonomous driving and roadside applications, roadside LiDAR data processing methods are primarily based on the background filtering method. The first step is to separate foreground and background; The second step is to cluster the moving points into vehicles or non-vehicles. Then tracked road users are used for speed estimation and safety analysis. Similar to autonomous driving LiDAR, the roadside LiDAR methods could also be categorized according to different ways of data representations, including point-based, voxel-based, projection-based, and spherical angular-based.

- **Point-based method**: Xiao et al. [9] developed Nearest-point methods to subtract background points based on the assumptions that background points have more neighbors than moving targets within a time window. Zhang et al. [10] developed roadside vehicle detection and tracking method based on ground plane removal. The data processing steps include the selection of the region of interest, ground plane point removal, vehicle clustering, and tracking. The data association method (DA) [11] preserved a pure background frame and identified background points in new frames with a pre-defined threshold (T) compared to all points in the reference frame.
- **Voxel-based method**: The 3D-density-statistic-filtering (3D-DSF) method in [12] was developed to track vehicles using roadside LiDAR sensors for Connected Vehicle applications. The algorithms include background filtering, lane identification, and vehicle speed tracking. In the background filtering step, the 3D space is divided into multiple cubes to estimate point density. The tracking step uses the average point as a tracking point to represent the detected vehicle. Rasterization is another technique used for voxel-based background modeling [13], which stores the voxel cube into an array format. Zhao et al. [14] implemented the 3D-DSF voxel-based method to detect and track pedestrians and vehicles at intersections. The performance of their proposed model is impacted by the density of points, occlusions, and perspective shadow.
- **Projection-based method**: This method is to reduce the 3D LiDAR to a 2D plane to leverage image-based modeling methods at the cost of losing 3D information. Zhang et al. [15] developed an image-based vehicle tracking method from roadside LiDAR after converting the 3D data into 2D images. To retrieve the transformation parameters, image registration is exploited.
- **Spherical Angular-based method**: Lee and Coifman [16] developed a vehicle detection and classification method, assuming that background's range at a given angle is constant. The background range is set as the median value at each angle as observed over multiple frames. Zhang et al. [17] proposed an automatic background construction and object detection method for roadside LiDAR data. The background dataset is constructed with the farthest distance in each horizontal-vertical angular value. By comparing the background dataset with new data according to the same horizontal and vertical angular value, object points were extracted and clustered to pedestrian and vehicle detection. [18] created an azimuth-height background filtering method using an Azimuth-Height table. The Azimuth-Height background filtering method compares the height of each point with the height of backgrounds.

Some other researchers advanced this area by integrating roadside LiDAR into traffic management. Zhao et al. [19] researched lane and movement-based traffic volume data collection using infrastructure-based LiDAR under different congestion levels and traffic compositions, covering signalized intersections, pedestrian crossings, work zones, stop-sign intersections, metered/unmetered ramps, and rural highways. Lv et al. [20] proposed a LiDAR-enhanced connected infrastructure solution to collect traffic data of traffic participants using roadside LiDAR and broadcast the message through DSRC to enable connected vehicle application. [21] used LiDAR detected vehicle trajectories to generate connected vehicle messages through the roadside unit to help connected vehicle applications.

The aforementioned methods generally use basic summaries from aggregated frames, such as maximum value, mean value, distance threshold, gradients, or density, to filter out backgrounds. Most methods lack transferability as parameters mainly rely on engineers' experience. For instance, the size of the 3D cube has a significant influence on detection performance, but cube size is hard to calibrate. Smaller cube sizes will significantly increase computational loads; larger cube size lowers the accuracy. Given the limitations of the earlier methods, a robust and intelligent LiDAR background modeling method needs to be developed. We developed advanced techniques through pattern decomposition and dynamic clustering with scalability and easy maintenance to address the challenges.

**Background Modeling**
The background modeling method is the first step of many video surveillance applications to understand video sequences. Each video frame is compared with the background video model to identify foreground objects with precise localization information. Bouwmans [22] provides a comprehensive survey paper. The background modeling method has three main steps: 1. background initialization using first N-frames; 2. Classification of pixels into foreground and background; 3. Background model maintenance over time. The paper also identified 13 challenging situations for background modeling: Noisy image; Camera jitter; automatic Camera adjustment; Illumination changes; Bootstrapping; Camouflage; Foreground aperture; Moved background objects; Inserted background objects; Dynamic backgrounds;

Beginning moving object; Sleeping foreground object; Shadows. The paper [23] classified the background modeling method into the following categories: Basic Background Modeling; Statistical Background Modeling; Fuzzy Background Modeling; Background Clustering; Neural Network Background Modeling; Wavelet Background Modeling; Background Estimation. The basic model uses mean [24], median [25], or histogram [26] to describe background pixels. The statistical model uses statistical variables such as Gaussian distribution [27, 28] Kernel Density Estimation [29] to classify pixels. The fuzzy background model [30, 31] uses a fuzzy running average or Type-2 fuzzy mixture of Gaussian. The background clustering model uses the K-means algorithm [32] or Codebook [33]. The neural network background modeling [34, 35] trains a set of weights on N clean background frames. The wavelet background model uses discrete wavelet transformation (DWT) [36]. The background estimation model is estimated with a filter such as a Wiener filter [37], a Kalman Filter [38], or a Tchebyche Filter [39]. Goyal and Singhai [40] reviewed several Gaussian Mixture Model for background/foreground detection, conducted comparative analysis, and analyzed the scope to improve them.

## METHODOLOGY

In this section, two data-driven algorithms, **D**ynamic **M**ode **D**ecomposition (DMD) and **C**oarse **F**ine **T**riangle **A**lgorithm (CFTA), were applied for roadside Lidar moving object detection and tracking. The first step of using the background modeling methods is to transform and reorganize LiDAR data from a packet of (X, Y, Z, intensity) to Azimuth, Range, and Intensity matrices.

### Data Transformation

The LiDAR sensor records distance relative to itself and intensity values (depending on the reflectivity of the object and the wavelength used by the LiDAR). Two types of packets were created, data packets and position packets. The position packets are referred to as GPS packets, and the data packets contain distance and intensity information. The LiDAR system used a spherical coordinate system initially, and then the spherical-format data are transformed in the format of XYZ coordinate. The 3D coordinates of each point, in the Cartesian and Spherical coordinates, are calculated as:

$$\begin{cases} x = r * cos(\omega) * cos(\alpha) \\ y = r * cos(\omega) * sin(\alpha) \\ z = r * sin(\omega) \end{cases} \quad (1)$$

where $r$ is the measured distance, $\alpha$ is the yaw angle around Z-axis and $\omega$ is the fixed pitch angle of each laser emitter.

The testing Velodyne LiDAR sensor generates one data frame after completing a 360° scanning, with the theoretical azimuth resolution of 0.18°. However, the azimuth angles between two emits often vary and deviate from the theoretical resolution. In this step, the original LiDAR data packet will be arranged into 1800 grids, based on an azimuth resolution of 0.2°. The range of azimuth α is from $-179.99° \sim 179.99°$, which needs to convert to $0°\sim359.99°$. A hash function rearranges the LiDAR point to its corresponding grid.

$$h(\alpha) = \mathrm{mod}\left(\left\lfloor \frac{\alpha}{0.02°} \right\rfloor + 1,\ 1800\right) \qquad (2)$$

if $h(\alpha)$ has collision, we will compare the range between two points that crash into the same azimuth grid. The smaller-range point will be preserved because the background points are often farther than foreground objects, and we want to keep the most informative data.

As shown in Figure 1, the LiDAR streaming data in .pcap file format are stored in Cartesian coordinates with additional information of intensity value. The number of beams and the elevation of each LiDAR beam are fixed for the LiDAR device. Therefore, we regard the elevation value as a known factor and don't need to keep a set of matrices for elevation data.

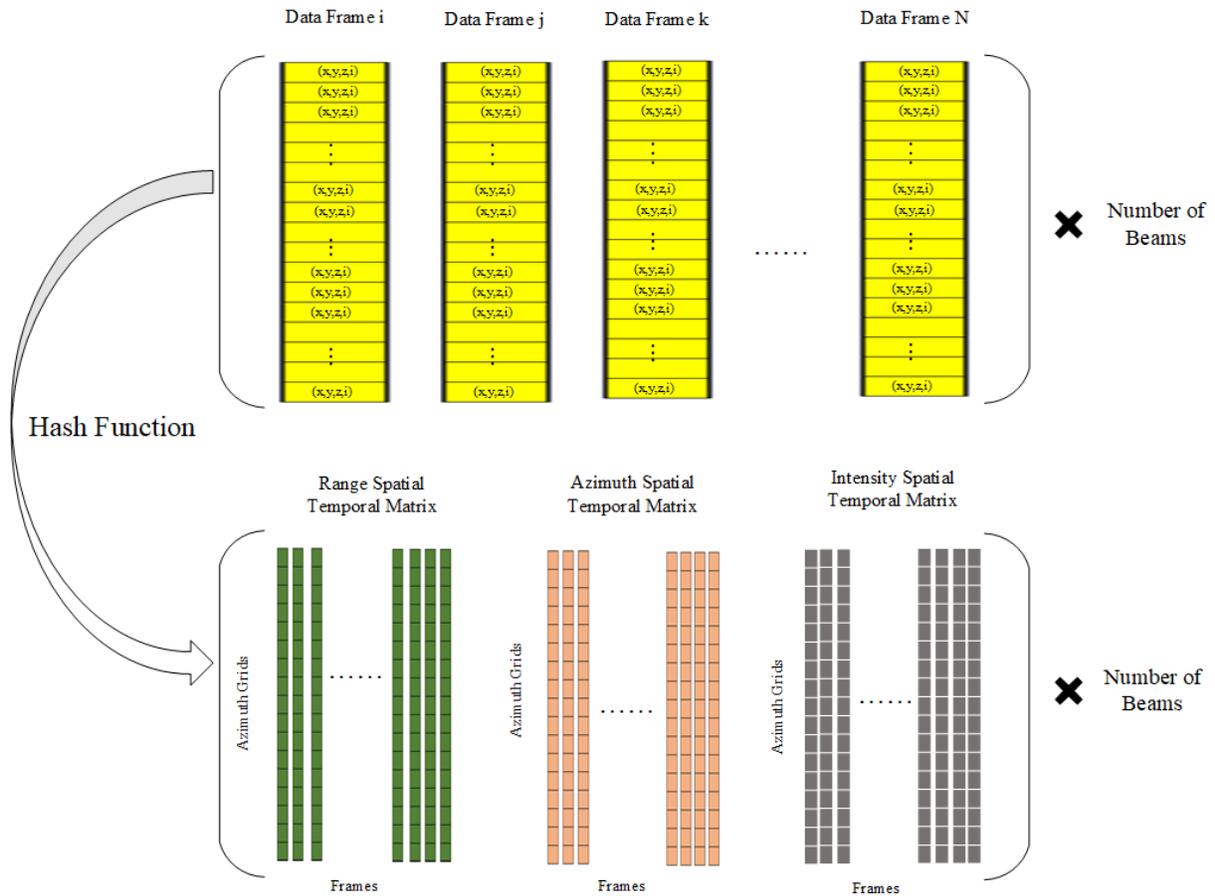

Figure 1 Transforming LiDAR Point Cloud into Azimuth Unit Spatial-Temporal Matrix

**Dynamic Mode Decomposition**

Dynamic mode decomposition (DMD) is a data-driven technique for discovering underlying patterns from high-dimensional data. It was first defined by Schmid and Sesterhenn [41, 42] to extract dynamic information from flow fields that can describe the physical mechanisms captured in the data sequence. The DMD methods are connected to the mathematical foundation that is readily interpretable using standard dynamic system techniques.

The goal of the DMD method is to extract the background mode for each channel of LiDAR. Then we will use the background mode to match the background points and filter out moving objects.

The LiDAR data at each beam can be considered as one slice of environmental information for each spin. The intensity data of $i_{x+1}$ at frame $(x + 1)$ is assumed to relate to previous intensity measurement of $i_x$ by linear operator A., the linear operator A is a time-independent operator that reflects the time evolution of each beam's intensity value.

$$\mathbf{i}_{x+1} = A \, \mathbf{i}_x \qquad (3)$$

The DMD algorithm is a regression method to estimate A that can characterize the intensity changes captured by each frame. The problem is formulated as follows:

$$I = \begin{bmatrix} | & \cdots & | \\ \mathbf{i}_1 & \ddots & \mathbf{i}_{m-1} \\ | & \cdots & | \end{bmatrix}, \quad I' = \begin{bmatrix} | & \cdots & | \\ \mathbf{i}_2 & \ddots & \mathbf{i}_m \\ | & \cdots & | \end{bmatrix} \qquad (4)$$

Where I is called the left intensity matrix, I' is called the right intensity matrix. I' has one frame difference compared to I. I' represents the time evolution of matrix I. The DMD algorithm seeks to find the best fit between the two matrices I and I' using a linear operator A.

$$I' = AI \qquad (5)$$

In order to solve A, the problem is converted to the following least-square problem.

$$\widehat{A} = \underset{A}{\operatorname{argmin}} \| I' - AI \|_F^2 \qquad (6)$$

By using the Moore-Penrose pseudoinverse, we obtain the estimator $\widehat{A}$:

$$\widehat{A} = I' I^\dagger \qquad (7)$$

The DMD mode that contains intensity information is the eigenvector of $\widehat{A}$. And each DMD mode corresponds to an eigenvalue of $\widehat{A}$. By finding the eigenvectors and eigenvalues of the matrix $\widehat{A}$, we obtain the DMD mode $\Phi = W$.

$$\widehat{A} W = W \Lambda \qquad (8)$$

The column of W are eigenvectors comprising of the dominant mode $\emptyset_j$ and $\Lambda$ is the diagonal matrix of eigenvalues $\lambda_j$. The Spatial-temporal Intensity Matrix be reconstructed using first $k^{th}$ modes, where $k \leq \min(n, m)$.

ST Intensity Matrix $\approx \Phi B \mathcal{V} =$

$$\underbrace{\begin{bmatrix} \emptyset_{11} & \cdots & \emptyset_{1k} \\ \vdots & \ddots & \vdots \\ \emptyset_{n1} & \cdots & \emptyset_{nk} \end{bmatrix}}_{\text{modes}} \underbrace{\begin{bmatrix} b_1 & \cdots & 0 \\ \vdots & \ddots & \vdots \\ 0 & \cdots & b_k \end{bmatrix}}_{\text{amplitudes}} \underbrace{\begin{bmatrix} 1 & \lambda_1 & \cdots & \lambda_1^{m-1} \\ 1 & \vdots & \ddots & \vdots \\ 1 & \lambda_k & \cdots & \lambda_k^{m-1} \end{bmatrix}}_{\text{dynamics}} \qquad (9)$$

Where $\Phi$ are dominant modes from the spatial-temporal map, matrix B is the matrix of amplitudes. $\mathcal{V}$ is the Vandermonde matrix representing the time evolution of DMD modes.

An intensity measurement $\mathbf{i}_t$ at frame $t \in 1, \ldots, m$ can be estimated as follows:

$$\tilde{i_t} = \sum_{j=1}^{k} b_j \emptyset_j \lambda_j^{t-1} \qquad (10)$$

Where $b_j$ is amplitude, $\emptyset_j$ is each DMD mode, and $\lambda_j^{t-1}$ is the time evolution of each intensity mode.

Let $t = 1$, and we obtain the following equation.

$$\tilde{i_1} = \sum_{j=1}^{k} b_j \emptyset_j \qquad (11)$$

So that the matrix B can be estimated as a least-square problem using the first scanline $i_1$ as the initial state.

$$\tilde{B} = \underset{B}{\mathrm{argmin}} \, \|i_1 - \Phi B\| \qquad (12)$$

Any DMD mode that does not change in time will have $\lambda_j = 1$, which forms the background of the intensity diagram.

In the intensity diagram, the intensity values of the background are highly correlated from one column vector to the next, suggesting the low-rank structure. The DMD algorithm separates background and foreground by decomposing the intensity diagram into low-rank (background) and sparse (foreground) components [43].

$$I_{DMD} = \text{background} + \text{forground} = \sum_p b_p \phi_p \lambda_p^{t-1} + \sum_{j \neq p} b_j \phi_j \lambda_j^{t-1} \qquad (13)$$

Where $|\lambda_p| = 1$. $t \in 1, \ldots, m$ is the data frame sequence.

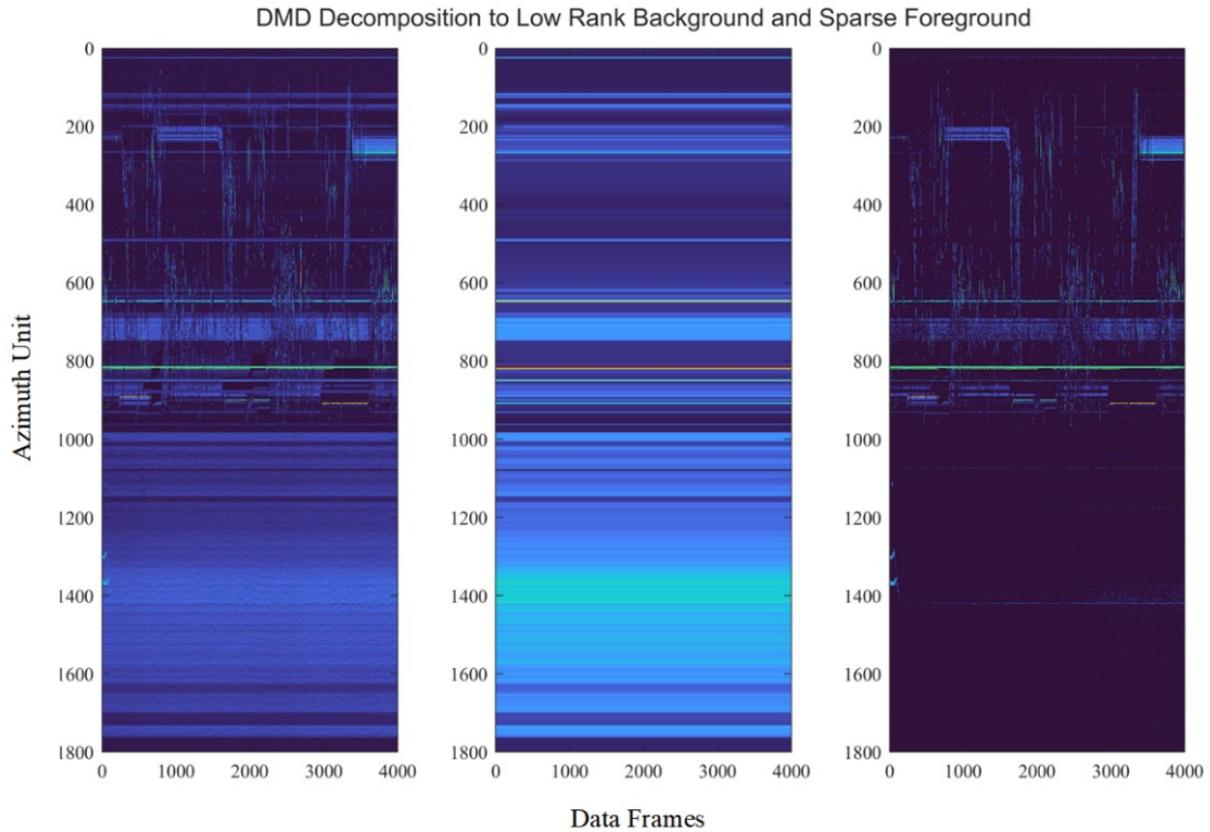

**Figure 2 DMD Decompose Spatial-Temporal Intensity Map into Low-Rank Background and Sparse Foreground**

The separation results are shown in Figure 2. The y-axis shows azimuth units of the LiDAR beam, and the x-axis is the accumulated data frame. The left figure is the original intensity image, the middle image is the background that is time-independent, and the right figure is the foreground moving objects. After obtaining the background intensity modes for all channels, we can use the background intensity value as a filter to detect the moving objects.

The DMD method will be applied to each beam to build a background filter, which will be used to separate moving objects from background objects.

*Algorithm: Compute DMD Background Modes of Intensity*
================================================================================
Input: Intensity Channel Azimuth Unit Diagram
Outputs: Background Intensity Mode for Each LiDAR Beam

--------------------------------------------------------------------------------
**For** Every Beam Intensity spatial-temporal matrix I
1. Calculate operator that fits between the following two matrices using Moore-Penrose pseudoinverse: $I' = AI$
   $\rightarrow A \approx I'I^{\dagger}$
2. Take SVD of I: $I \approx U\Sigma V^*$
3. Reduced matrix and obtain of $\tilde{A}$ by projecting A onto $U_r$: $\tilde{A} = U_r^* A U_r = U_r^* X' V_r \Sigma_r^{-1}$
4. Eigen decomposition: $\tilde{A}W = W\Lambda_r$
5. Compute modes: $\Phi = X'V\Sigma^{-1}W$
6. Obtain the low-rank background mode $\phi_p$, whose corresponding eigenvalue is asymptotically close to 0.

**End For**

## Coarse-Fine Triangle Algorithm (CFTA)

The Triangle algorithm is a dynamic clustering method based on histogram analysis, considering that the static infrastructure backgrounds are the farthest objects hit by the laser beams. The first step is to construct a histogram of ranges vs. frequency for all elevation-azimuth units. Figure 3 is generated after accumulating 4000 frames using testing LiDAR, and the range information from Beam ID 90 at Azimuth Grid 80 was randomly selected. Like in the histogram, the background range value is around 19 meters, while the foreground moving object's ranges are about 15 meters. The background/foreground ratio is about 7:1 under heavy traffic conditions. Draw a line between the highest range value of the histogram $h_{max}$ and the minimum range histogram $h_{min}$ (for LiDAR data, the minimum histogram is by default at a distance 0 because the emitted laser points either hit an object and returned with a positive range value or never return). Then the algorithm will calculate the point to line distance d and increase $h_{min}$ and repeat for all histogram h until $h = h_{max}$. The threshold value becomes the bin edge of range value for which the distance d is maximized. For the triangle algorithms, the only concern is the bin size of the histogram to make sure the background counts will fall into the same bin, at the same time separable from moving objects. We developed a Coarse-Fine Triangle Algorithm (CFTA) to automate the bin size selection process. For the coarse step, the algorithm estimates where the highest peak is. In the fine step, outliers were removed, and a new bin size was determined to find the threshold value.

The triangle thresholding aims to classify point clouds into either background points or moving objects based on the range information. The CFTA can automatically select the threshold range values for the moving vehicle detection. The method is developed on two assumptions: 1. The static background objects occupied most of the frequencies in the LiDAR point clouds. 2. The background points have the farthest distance and are distributed normally with standard measurement errors. This technique is particularly effective when the background objects point clouds produce a dominant peak in the histogram [44]. Figure 4(A) illustrates the situation that the laser beam reached the farthest object in the environment, which often becomes the background. Figure 4(B) describes the situation that laser beam hit on a passenger car, which is our target for the segmentation task.

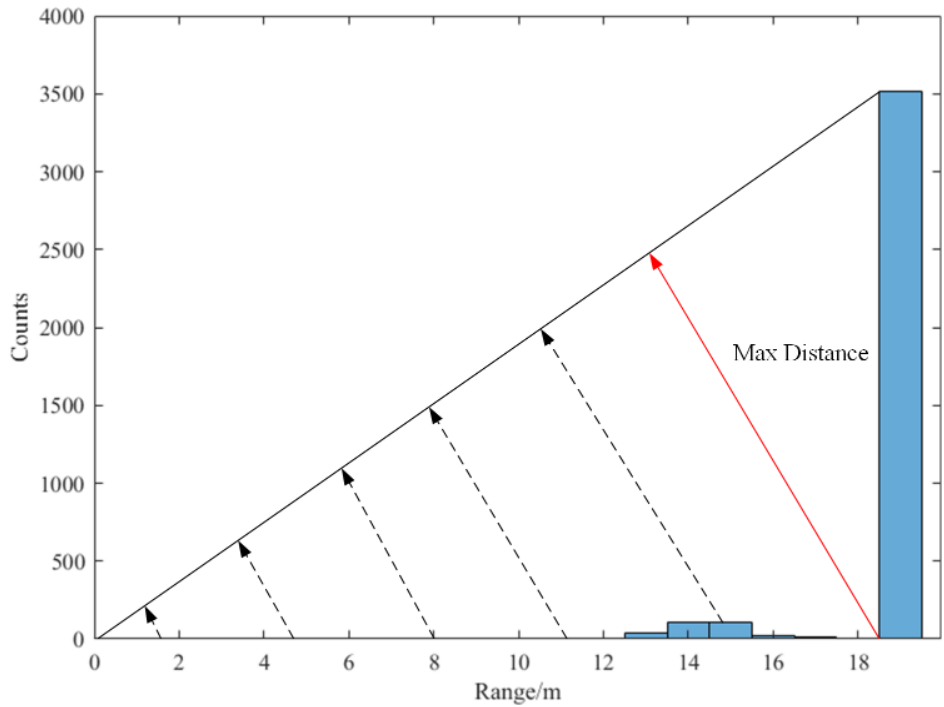

Figure 3 The triangle algorithm finds the threshold value that maximizes the distance $d$

*Algorithm: Coarse-Fine Triangle Algorithm*
========================================================================================

Input: Aggregated Distances for Elevation-Azimuth Angular Units
Outputs: Background vs. Foreground Thresholds

----------------------------------------------------------------------------------------

**For** Every Elevation and Azimuth Unit

    1. Nonreturnable points:
- If non-returnable points are the majority, then consider a maximum range distance. (e.g., 200 meters) as backgrounds.
- Else, remove all non-returnable points.

    2. Coarse Step:
- Use the default function to generate the histogram counts.
- Find the bin edge value $B_{max}$ that contains maximum counts
- Delete the points larger than $B_{max} + 2\sigma$, where $\sigma$ is the standard deviation

    3. Fine Step:
- Find the maximum range value $R_{max}$ after Coarse Step
- New bin_size = $R_{max}/100$

    4. Create new histogram counts on $[0, R_{max}]$ with new bin size.
    5. Apply the Triangle Algorithm in Figure 3.
    6. Save the threshold for this Elevation-Azimuth Angular Unit

**End For**

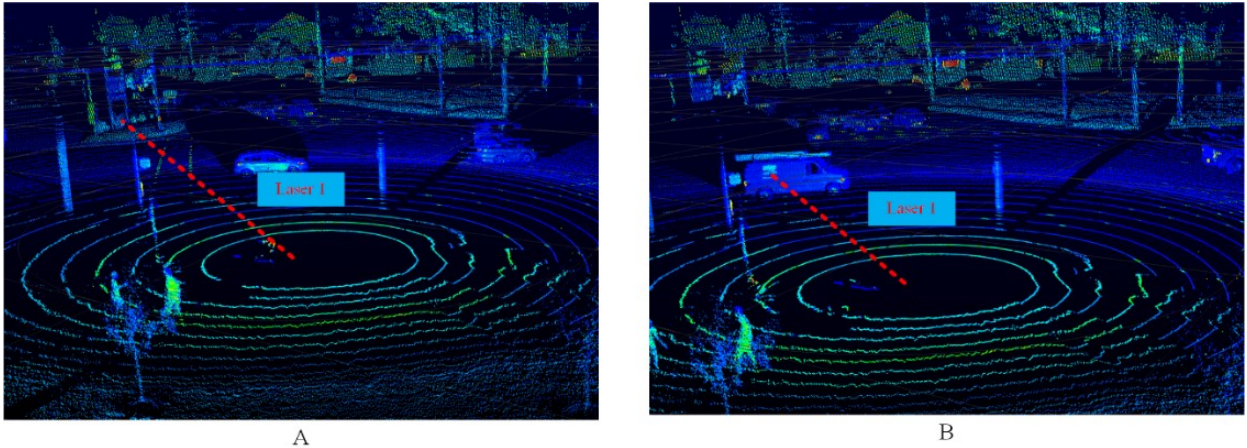

Figure 4 The LiDAR Laser Beam was Intercepted by Moving Vehicle

**EXPERIMENT DESIGN**

This data collection is to test Intelligent Transportation System Infrastructure at intersections through immediate data collection and analysis capability. This project aims to test innovative sensing and detection methods to evaluate and monitor signal performance in real-time. The outcome of the project will be used to improve intersection safety, reduce congestion, and have an environmental impact. The testing site is selected from a key New Jersey arterial corridor on Oct 20, 2021, from 3-6 pm at US1 at Bakers Basin. The 3-hour data include high-resolution GoPro video, 128 beam Velodyne Alpha Prime LiDAR data, connected vehicle SPaT, and MAP data. The camera was mounted on the roadside pole, and the LiDAR sensor was mounted with a tripod at the walkway, powered by high-capacity batteries and a solar panel. The GoPro generated 80G data at 60 frames per second during the three-hour periods, and the Alpha Prime Velodyne LiDAR generated 70G data at 10 Hz. The LiDAR has similar data storage efficiency and somewhat reduced the data size compared to video data.

In Figure 5, the experiment setup was displayed to show the coverage for video and LiDAR detection at the same timestamp. The vehicle detection and tracking are processed using Yolov5 and DeepSort for comparative analysis. The LiDAR Sensor was only installed at the height of 1.7 meters and already can sufficiently provide a wide range of coverage and holistic 3D measurements of the surrounding environment. Although the camera is installed at a height of about 5 meters, it is still difficult to cover the entire intersection area from all directions. Compared to the camera detector, the LiDAR sensor shows excellent potential and will play a significant role as an intelligent infrastructure solution in the following years.

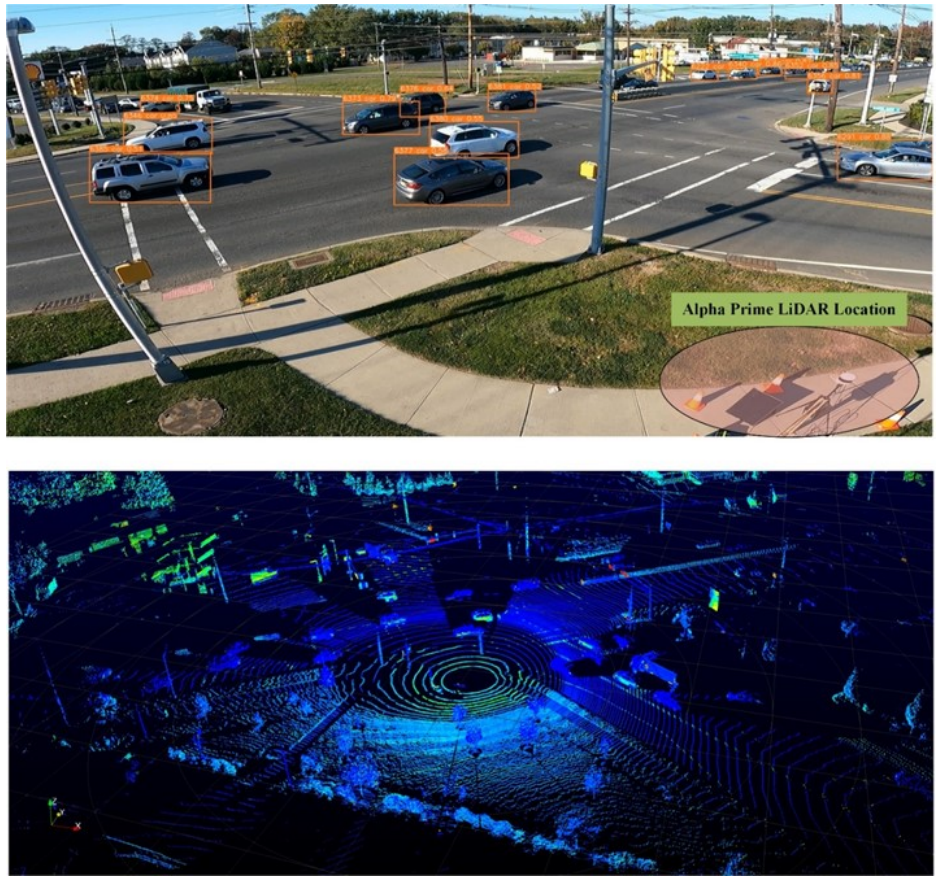

**Figure 5 Testing Site for Traffic Data Collection Using Camera and LiDAR Sensor**

## MODEL EVALUATION

In this session, we will break the entire solution into sequential steps and examine the model results in detail. The overall workflow is shown in Figure 6. The ROI filter, noise removal, clustering, bounding box detector are considered general approaches. The tracking module was implemented with fine-tuned parameters from off-the-shelf packages. The two new algorithms, including DMD intensity background subtraction and CFTA for range background subtraction, are integrated as one module. The extracted vehicle movements can be applied to many mobility or safety applications. For example, the vehicle counts for each turning movement could be used for signal optimization to assess whether the phase split is efficient.

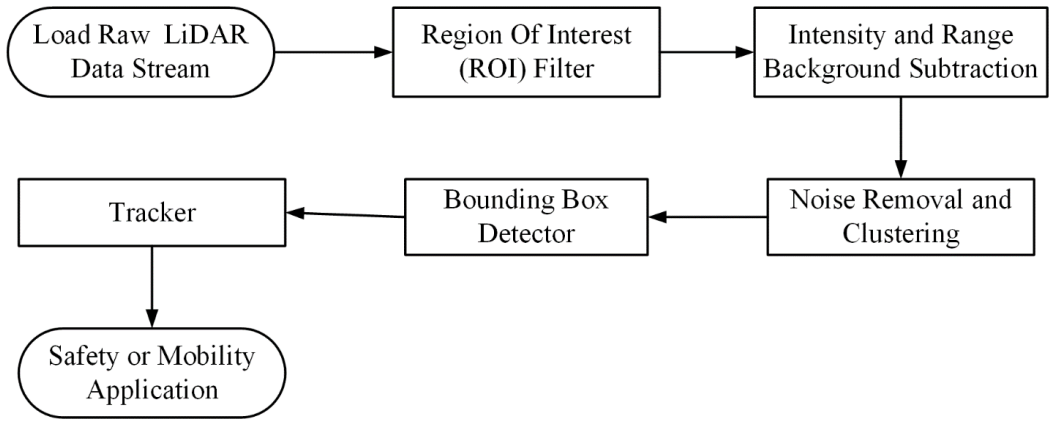

Figure 6 Workflow of Infrastructure LiDAR Background Subtraction

**ROI Filter**
As the infrastructure LiDAR is static, accurate GPS coordinates can be obtained in practice. Therefore, the non-drivable space within the monitored area could be removed using geofencing methods. In Figure 7, the raw LiDAR data are filtered by projecting all points into the X-o-Y plane using a binary mask.

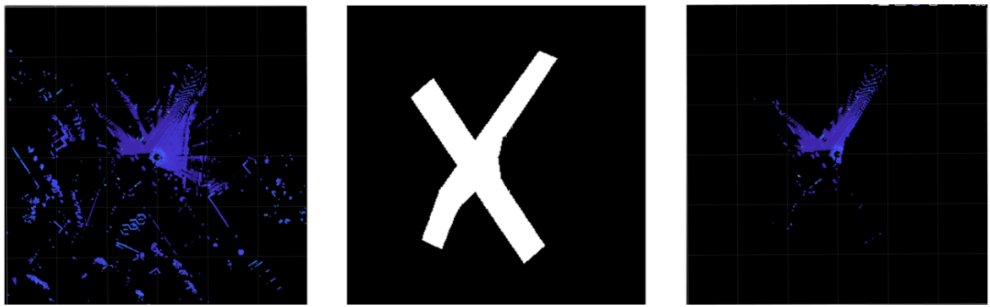

Figure 7 Region of Interest (ROI) filter

**Background Subtraction**
The background subtraction methods are directly performed on spherical coordinates, which are the original coordinates of data collected from the LiDAR sensor. Therefore, it would save the computation from converting spherical data to Cartesian coordinates within the sensor chip. The mainstream LiDAR processing methods are based on the cartesian coordinates, and the data are saved in sparse matrices. With efficient background subtraction, more than 90% of data can be eliminated. Therefore, using the spherical coordinates system could significantly improve the LiDAR point cloud acquisition and transmission efficiency.

**Noise Removal and Clustering**
The noise removal is based on the Local Outlier Factor (LOF) Algorithm. A point will be considered noise if its K-distance, the distance between the point and its $K^{th}$ nearest neighbor, is smaller than a threshold. The point cloud clustering step is also distance-based and segments all point cloud data into clusters and returns cluster labels of all cloud points.

**Bounding Box Detector and Tracking**

Upon finishing clustering, we then estimate the bounding box to each cluster that is greater than the minimum threshold number of points. Then the detected object is encoded into the state-space model that contains the objects' corresponding measurements and transition of state (speed in x, y z dimension, and turning rate). A joint probabilistic data association (JPDA) tracker with an interacting multiple model (IMM) filter is applied to update the tracked list of objects for each frame. In Figure 8, the model outputs after all steps are presented, showing the vehicle detection and tracking results from three phases of the signalized intersection. In the first column, the foreground moving vehicles are colored green, and the background LiDAR point clouds are colored in purple. The middle column pictures contain tracking module outputs, where the red boxes are detected objects from the detector module, and the green box is confirmed tracks with certain confidence with the tracking history.

We also presented the video detections at the same timestamp with LiDAR data using Yolov5, which was trained on the coco dataset and DeepSort for real-time vehicle detection and tracking. As you can see, the LiDAR sensor provides a broader field of view than the GoPro camera. For Phase B, the pre-trained deep learning model missed three vehicles passing the intersection due to the impact of the streetlight pole. However, the streetlight pole has little effect on LiDAR detection. The proposed LiDAR detection model showed excellent reliability and was comparable to one of the most advanced video detectors.

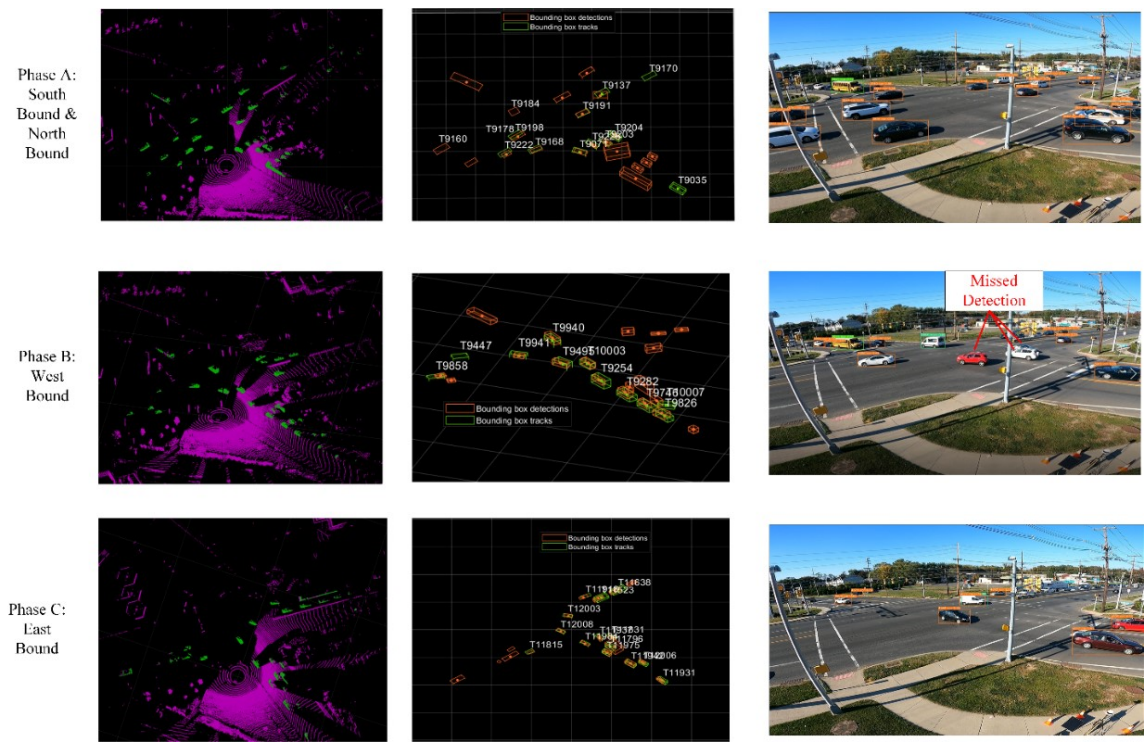

**Figure 8 Proposed LiDAR Object Detection and Tracking Compared to Deep Learning Video Detection**

**Path Level Evaluation**

Movement counting is an essential input for the signalized intersection to optimize the timing parameters. Figure 9 shows LiDAR detected vehicle trajectories grouped by traveling paths in different colors. The second half of the picture is vehicle detection and tracking results from a commercial AI data collection platform [45], which can generate a report of traffic counts at 15-minute intervals with more than 95% accuracy.

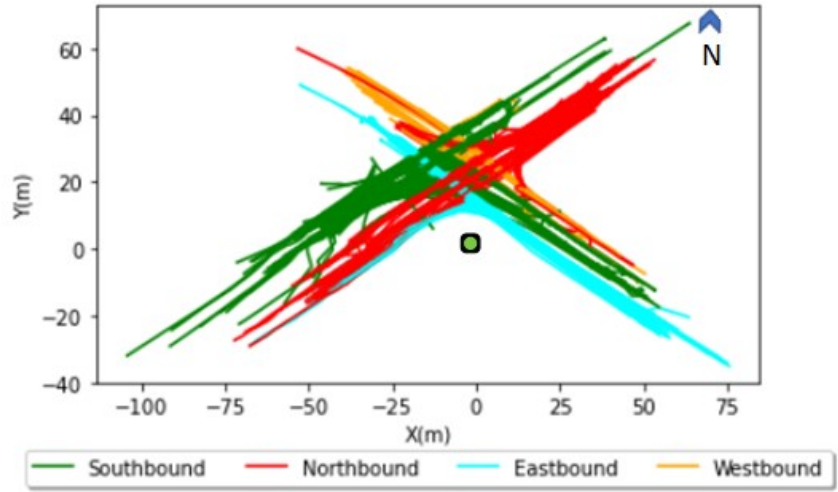

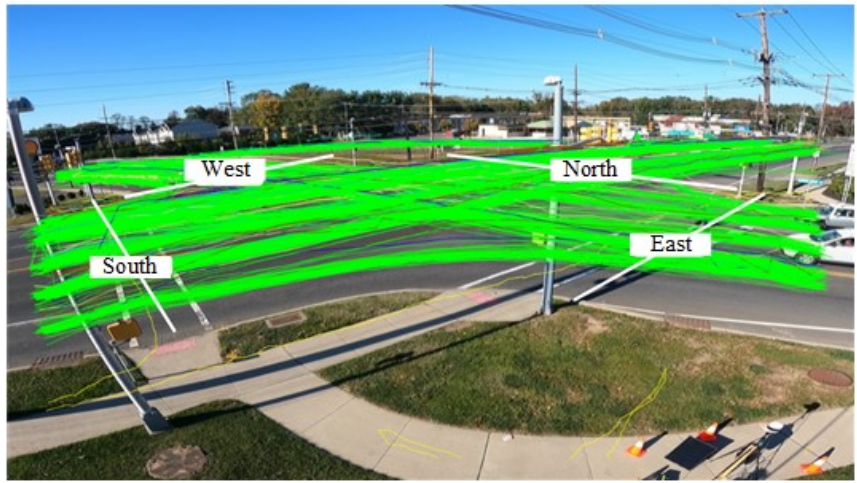

**Figure 9 LiDAR Detected Trajectory and Video Detected Trajectory as Benchmark**

**Table 1 Vehicle Movements Counting Evaluation**

|  | Total Count | Eastbound | Westbound | Northbound | Southbound |
|---|---|---|---|---|---|
| LiDAR | 1008 | 143 | 68 | 391 | 406 |
| Video Benchmark | 1064 | 125 | 78 | 448 | 414 |
| Error Rate | 5.26% | 14.40% | 12.82% | 12.72% | 1.93% |
| Accuracy | 94.74% | 85.60% | 87.18% | 87.28% | 98.07% |

In Table 1, vehicle movement countings for all four directions are presented. The overall counting accuracy is 94.74%. As roadside LiDAR traffic detection is still an emerging application, no previous research has been done by validating roadside LiDAR traffic detection with a commercial AI-based video detection platform. This movement count assesses the background subtraction performances and also tests the detection and tracking modules. The main reason that causes counting errors is that the vehicles in the farther lane to the LiDAR sensors are often blocked by nearby vehicles from other paths. The blind zones on the LiDAR point clouds pose significant challenges to the tracking modules due to the hit-and-miss vehicle presences and partial occlusions.

**Point Level Evaluation**

The following Table 2 was concerned with segmentation results at point level compared to the SOTA method [17]. In the training process, the baseline model accumulated 2000 frames and applied mean and max values to remove foreground points. The preserved background points are stored and used to judge whether new points are moving objects by comparing them with reference points. We then randomly selected data frames and manually processed background removal to generate ground truth data. By comparing model-filtered points with ground truth points, we can say if the detected point is True Positive, False Positive, or False Negative. We created three detection evaluations using the common classification metrics, Precision, Recall, and F1 Scores. The precision score tells us what percentage of detected points is correct. The recall score tells us what the percentage of detected foreground points is. The $F1$ score is a harmonic mean of precision and recall scores to balance two measurements.

Table 2   Point Level Evaluation at Different Ranges

|  | [0, 30) Meter Range | | [30, 100) Meter Range | |
|---|---|---|---|---|
|  | Proposed Method | Reference Model | Proposed Model | Reference Model |
| Precision | **99.23%** | 96.27% | **97.69** | 90.31% |
| Recall | 73.13 | **82.08%** | 70.08 | 67.87 |
| F1 Score | 84.23% | **88.61%** | 81.61 | 77.50% |

It can be seen from Table 2 that our new methods have the best precision on both [0, 30) and [30, 100) meter ranges. The high precision and relatively low recall scores suggest that our model is more rigorous than the baseline model. Overall, our approach surpassed the SOTA baseline on 4/6 evaluation categories. The reference model takes 2.98 seconds to process one frame since it compares new data points with accumulated background points. Our method runs at 0.26 seconds per frame, which is faster than baseline models. Figure 10(A) is the manually processed ground truth data; Figure 10 (B) and (C) are our model detection results and baseline model detection results. From Figure 10, we can see that the reference model tends to preserve more background points than our proposed methods.

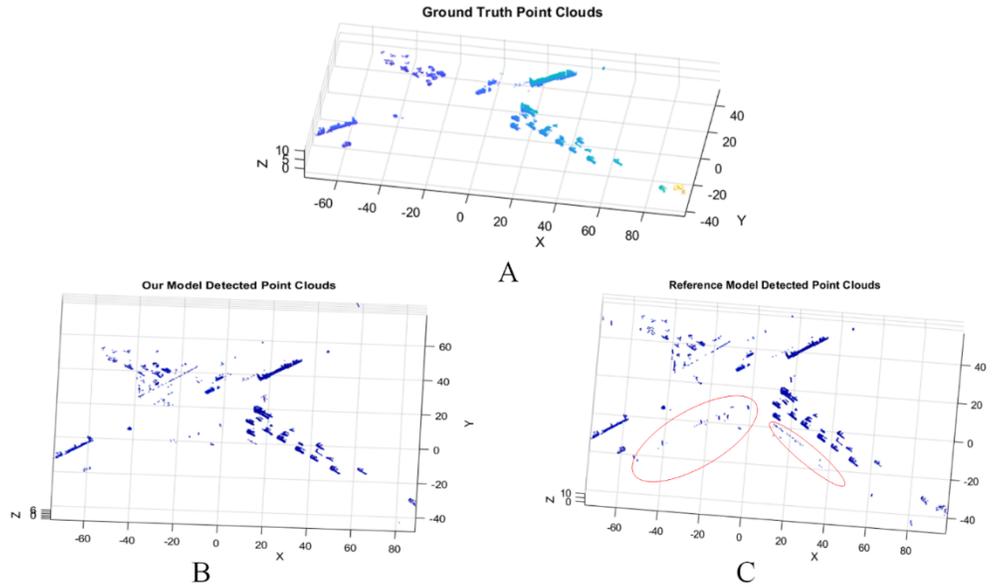

**Figure 10 Point Level Evaluation Comparison (Red Cycle: Remaining Backgrounds)**

## DISCUSSIONS

The roadside LiDAR data has different characteristics than mobile LiDAR data. First, most of the point clouds in the roadside LiDAR model are static background, while the mobile LiDAR point clouds model contains mainly the changing environment; Second, with the increasing distance between road users and the LiDAR sensor, the gaps among laser beams get more significant, resulting in the more unseen area and fewer rings on detected objects. Third, the roadside LiDAR is usually installed at an elevated location to monitor a large area, while autonomous driving LiDAR mainly scans side-by-side vehicles. Figure 11 presents the experimental results using PointPillars [3] deep neural networks trained on PandaSet [46], which contains 2560 preprocessed organized LiDAR scans of various driving scenes. The data set provides 3-D bounding box labels for different object classes, including car, truck, and pedestrian. As shown in Figure 11, the PointPillars attains effective results on its mobile LiDAR dataset (Table 3). When the confidence threshold was set as the default value of 0.5, the pre-trained deep learning model generates zero detections, which is not applicable to our roadside LiDAR purpose (Figure 11(B)). After lowering the confidence level to 0.3, the pre-trained model only gives a few disoriented detection results (Figure 11(C)). The autonomous driving LiDAR training data are generated with 64 beam devices, while the roadside LiDAR has 128 beams. Another observation is that autonomous driving LiDAR datasets are biased on nearby vehicles with the same traveling direction, while roadside LiDAR dataset contains vehicles coming from any direction.

Table 3 Testing Performance on Mobile LiDAR Data vs. Roadside LiDAR

|  | PandaSet LiDAR Dataset | | Roadside LiDAR | |
|---|---|---|---|---|
|  | Precision | Recall | Precision | Recall |
| Car | 90.03% | 52.99% | NA | NA |
| Truck | 36.76% | 58.59% | NA | NA |

Note: NA stands for Not Applicable

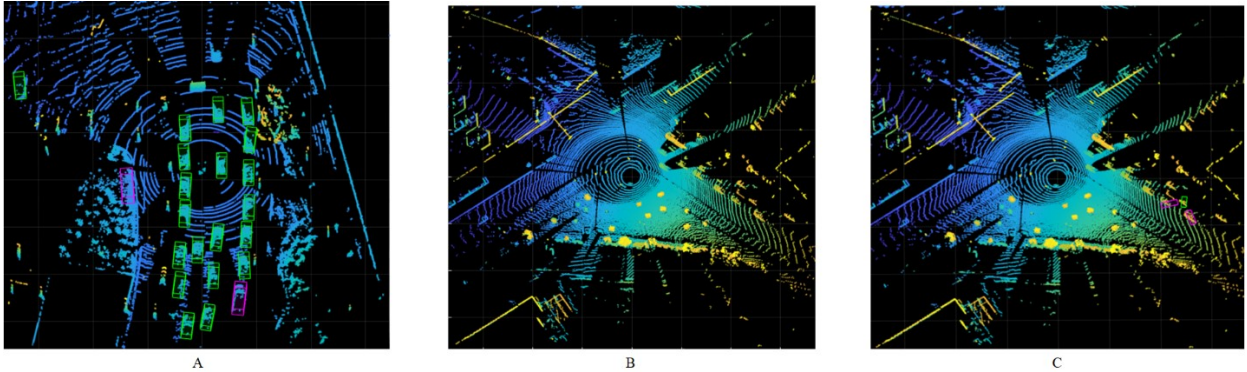

Figure 11 Trained Deep Neural Network on Self-driving LiDAR and Roadside LiDAR Data (A. PointPillars Model on Mobile LiDAR Data; B. PointPillars Model on Roadside LiDAR Data with Default Confidence = 0.5; C. PointPillars Model on Roadside LiDAR Data with Confidence = 0.3; Magenta: Truck; Green: Car )

The high-resolution point cloud data will support the next-generation research on 3D big data sensing and analytics by creating the digital twin of infrastructure systems in a holistic 3D environment. The roadside LiDAR object detection could be used to explore many underlying scientific problems, including transportation, infrastructure, energy, public service, and human activity systems and their interactions. To examine the model performance on different scenarios, the method was also implemented in an urban environment as part of the Middlesex County Smart Mobility Testing Ground (SMTG) to establish a living laboratory for smart mobility and smart city technology research. Our model was further tested at an intersection in downtown New Brunswick, New Jersey. The proposed method can adapt effortlessly to a new scenario with less than a couple of hours for preparation and re-calibration. Figure 12 shows vehicle detection results from the portable setting and the urban scenario with a permanent power supply and communication cables. The animated visualizations of over 1000 frames can be found in this project's public repository [47].

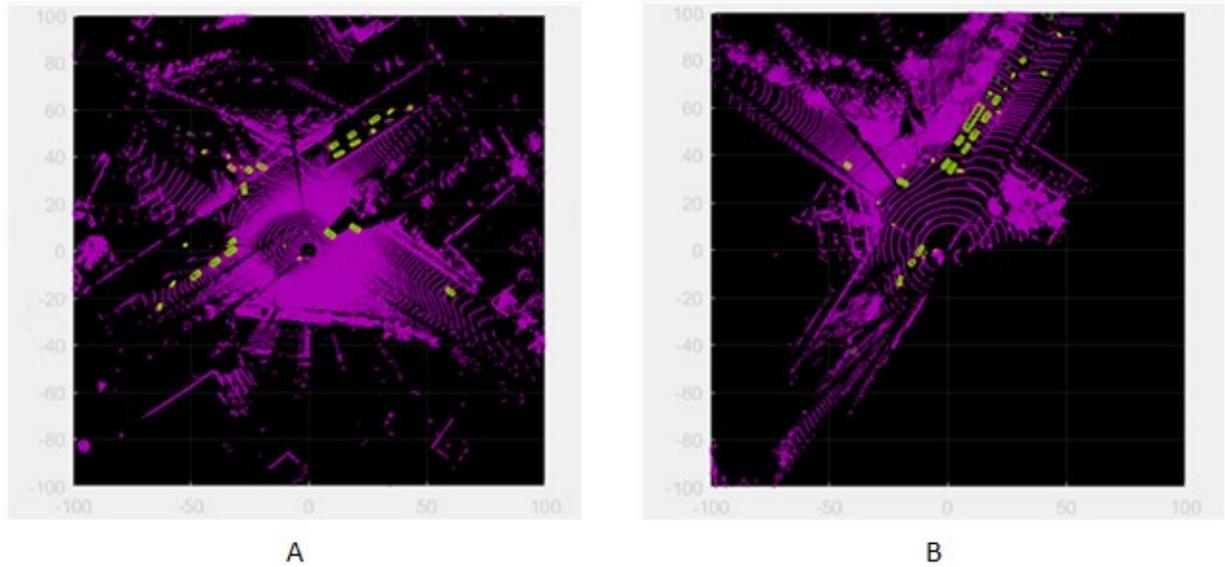

**Figure 12  Proposed Methods in Different Scenarios (A. Portable Tripod Installation at a Suburban Corridor, B. Permanent Installation at an Urban Intersection)**

**CONCLUDING REMARKS**

In this paper, we developed a novel background subtraction method with unsupervised learning algorithms for infrastructure LiDAR object detection and tracking. The main contributions of this paper to the existing works of literature are summarized as follows:

1. Our method integrates the range and intensity information for point cloud object detection, which can also be used independently. As a result, this method can reduce 90% redundant background points and increase the data acquisition efficiency.
2. Instead of converting the point clouds into 3D voxels, our methods transform the unstructured point clouds into structured representation to be processed by a 3D object detector with reduced dimensions. With proper data transformation, we bridge the gap between image-based background modeling and point clouds background modeling, making a rich body of well-studied image-based techniques suitable for LiDAR data.
3. The proposed methods are built on unsupervised learning that automatically discovers the structures from data. The two algorithms require very few parameters, which means more robust and easier auto-calibration and deployment. For intensity-based algorithms, the only parameter is the intensity threshold that differentiates sparse foregrounds from the low-rank background modes. The Coarse-Fine Triangle Algorithm is even better as a parameter-free algorithm.
4. Compared to the deep learning-based LiDAR object detection methods, our method shows more heuristics and better expandability. It does not need to collect large amounts of training data, sophisticated network design, and GPU to support the functionality.

5. Compared to the SOTA roadside LiDAR background modeling methods, our method runs faster with better performance, evidenced by point-level assessment. In addition, the proposed background model is easy to maintain.

LiDAR-based 3D object detection is a challenging task, which requires high detection performance and fast inference. The occlusions of LiDAR data are the main challenge of detection and tracking due to blind zones and increased monitoring area for roadside application, making the inherently sparse point clouds even more complicated to process. A potential solution is to stitch multiple point clouds together to increase data points or fuse information from different sensors.

## FUNDING STATEMENT

This work is funded by the New Jersey DOT Real-time signal performance measures (Project No. 2016-14); New Brunswick Innovation Hub Smart Mobility Testing Ground (SMTG) Contract Numbers: 21-60168.

## ACKNOWLEDGMENT:

The preprint version of this article could be found in [48].